\documentclass[runningheads]{llncs}
\usepackage{graphicx}

\usepackage[hang,flushmargin]{footmisc}
\usepackage[misc,geometry]{ifsym}
\usepackage{cite} 
\usepackage[T1]{fontenc}
\usepackage{amsmath}
\usepackage{mathrsfs}
\usepackage{graphicx}  
\usepackage{epstopdf}
\usepackage{bbm}
\usepackage{float}
\usepackage{multirow}
\usepackage{booktabs}
\usepackage{mathrsfs}
\usepackage{amssymb}
\usepackage[colorlinks,linkcolor=red]{hyperref}
\usepackage{siunitx}

\usepackage{arydshln}

\makeatletter

\newcommand{\Rmnum}[1]{\expandafter\@slowromancap\romannumeral #1@}
\makeatother
\usepackage{booktabs}
\usepackage{amsfonts,amssymb}
\usepackage{amsmath,bm}
\usepackage{array}
\newcommand{\PreserveBackslash}[1]{\let\temp=\\#1\let\\=\temp}
\newcolumntype{C}[1]{>{\PreserveBackslash\centering}p{#1}}
\newcolumntype{R}[1]{>{\PreserveBackslash\raggedleft}p{#1}}
\newcolumntype{L}[1]{>{\PreserveBackslash\raggedright}p{#1}}
\usepackage{algorithmicx,algorithm}
\makeatletter

\newcommand{\para}[1]{\vspace{.05in}\noindent\textbf{#1}}
\def\ie{\emph{i.e.}}

\usepackage{xcolor}

\begin{document}
\title{MuST: Multimodal Spatiotemporal Graph-Transformer for Hospital Readmission Prediction}
%
\titlerunning{Multimodal Spatiotemporal Graph-Transformer}
    
\author{
Yan Miao, Lequan Yu
}
\institute{
The University of Hong Kong, Hong Kong SAR, China\\
\email{ymiao7@connect.hku.hk lqyu@hku.hk} 
}

%
%

\maketitle 

\begin{abstract}

Hospital readmission prediction is considered an essential approach to decreasing readmission rates, which is a key factor in assessing the quality and efficacy of a healthcare system.
Previous studies have extensively utilized three primary modalities, namely electronic health records (EHR), medical images, and clinical notes, to predict hospital readmissions. 
However, the majority of these studies did not integrate information from all three modalities or utilize the spatiotemporal relationships present in the dataset. 
This study introduces a novel model called the Multimodal Spatiotemporal Graph-Transformer (MuST) for predicting hospital readmissions. 
By employing Graph Convolution Networks and temporal transformers, we can effectively capture spatial and temporal dependencies in EHR and chest radiographs. 
We then propose a fusion transformer to combine the spatiotemporal features from the two modalities mentioned above with the features from clinical notes extracted by a pre-trained, domain-specific transformer.
We assess the effectiveness of our methods using the latest publicly available dataset, MIMIC-IV. 
The experimental results indicate that the inclusion of multimodal features in MuST improves its performance in comparison to unimodal methods. 
Furthermore, our proposed pipeline outperforms the current leading methods in the prediction of hospital readmissions.
%


\end{abstract}

\section{Introduction}
%

Hospital readmission prediction is considered a crucial aspect of clinical decision-making. 
It refers to predicting the occurrence of a patient who has been discharged from a medical facility and is subsequently admitted to either the same institution or a different one within a designated timeframe, typically ranging from 30 to 90 days, subsequent to their prior visit \cite{wang2021predictive}. 
The concept of a revisit typically denotes an instance where a patient's previous in-patient visit was either insufficient or ineffective in addressing their medical needs \cite{wang2021predictive}. 
As a result, decreasing avoidable hospital readmissions has become a top priority for healthcare systems.

The advancement of deep-learning approaches has significantly contributed to the research on hospital readmission prediction in recent years \cite{huang2019clinicalbert, tang2022multimodal, wang2018predicting, reddy2018predicting, allam2019neural, zhang2020combining}.
Existing studies have approached the problem from various perspectives. 
\cite{allam2019neural, reddy2018predicting, wang2018predicting} extracted temporal relationships in longitudinal electronic health records (EHR) clinical data using CNN-, RNN-, and LSTM-based frameworks. 
%
%
Some works utilized clinical notes for hospital readmission prediction as they offer a more comprehensive understanding of the patient compared to structured features \cite{huang2019clinicalbert}. 
For instance, ClinicalBERT \cite{huang2019clinicalbert} is a variant of bidirectional encoder representations from transformers (BERT) \cite{devlin2018bert} model that is specifically pre-trained on clinical notes from the MIMIC-III dataset \cite{johnson2016mimic} and finetuned for the purpose of predicting 30-day hospital readmission. 
%
%
%
%
%
Meanwhile, several studies have attempted to predict readmission by aggregating the diverse perspectives of information provided by different data modalities within the hospitalization records of patients.
Zhang et al.\cite{zhang2020combining} proposed a multimodal CNN-based approach that combined structured EHR and unstructured clinical notes to predict hospital readmission. 
Recently, a novel pipeline called the multimodal, spatiotemporal graph neural network (MM-STGNN) \cite{tang2022multimodal} has been proposed to handle the task. 
It fused longitudinal chest radiographs and EHR of patients, and employed a graph-based approach to capture patient similarity. 
The pipeline's multimodal and spatiotemporal features resulted in better performance than traditional baselines like gradient-boosting and LSTM. 
However, the prior research works have a common limitation: they failed to integrate all the valuable EHR, medical imaging, and textual data modalities for better hospital readmission prediction.
Moreover, the late-stage multimodal fusion strategy in MM-STGNN is insufficient to model the fine-grained interaction of different modalities, as it only extracts a global representation for each modality.


%
%

%
%

%
%

The Transformer architecture \cite{vaswani2017attention}, an attention-based deep sequence model, has gained significant attention in the deep learning community due to its exceptional performance across various tasks \cite{devlin2018bert, parmar2018image, li2019visualbert, ramesh2021zero}, where most of them either worked under a unimodal \cite{devlin2018bert, parmar2018image} (i.e., either text or imaging) or visual-language scenario \cite{li2019visualbert, ramesh2021zero}. 
%
%
Recently, a few studies \cite{gabeur2020multi, prakash2021multi, nagrani2021attention} have discovered that the integration of multimodal data using transformers enhanced task performance, providing a more effective approach to modeling dependencies between different modalities. 
A transformer model treats each modality as a distinct input sequence and utilizes the attention mechanism to selectively attend to parts of the input sequences based on their relevance to the original task. 
This enables the model to capture intricate relationships between modalities and produce accurate predictions.
Previous studies have also explored the utilization of transformers in the context of AI in healthcare \cite{huang2019clinicalbert, li2020behrt, rao2022targeted}.
However, there are limited works to explore how to use transformers to integrate multimodal medical data, and we have yet to find any existing research that integrates multimodal features using transformers for the hospital readmission prediction task. 
%
%
%

In this study, we introduce a novel \underline{Mu}ltimodal \underline{S}patialtemporal Graph-\underline{T}ransformer framework (\ie MuST), which integrates three modalities — EHR, medical image, and clinical note — to collectively predict 30-day all-cause hospital readmission. 
Specifically, we construct a graph to capture the spatial dependencies between hospital admissions for EHR and imaging modalities and then adopt the pre-trained BioClinical BERT \cite{alsentzer2019publicly} to extract meaningful representations from the clinical notes. 
We further incorporate the temporal relationships of the EHR and medical images by processing the spatial features of each modality with two distinct temporal transformers.
Finally, a fusion transformer is employed to integrate the spatiotemporal features of the EHR and images with the informative representations of clinical notes. 
Extensive experiments were conducted on the publicly available MIMIC-IV \cite{Johnson2023, physiobank}, MIMIC-CXR-JPG \cite{johnson2019mimic, physiobank}, and MIMIC-IV-Note \cite{johnson2023mimicnotes, physiobank} datasets to validate the effectiveness and superiority of our proposed framework.

\begin{figure}[t]
\includegraphics[width=\textwidth]{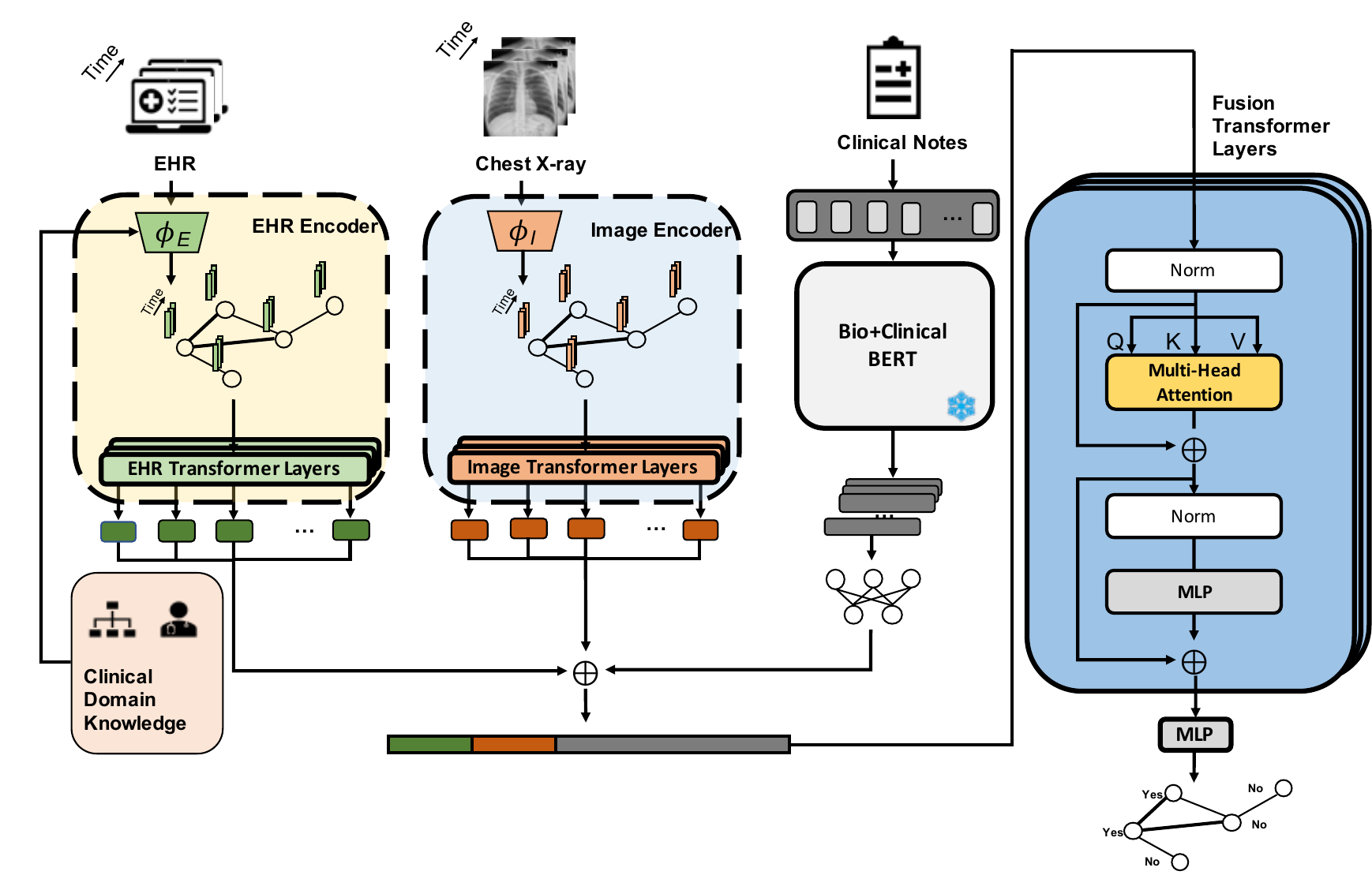}
\caption{Overview of the proposed MuST framework. The longitudinal EHR and chest radiographs undergo a spatiotemporal pipeline (i.e., \textbf{ST-Transformer}) consisting of a feature extractor, a relationship graph (each node represents a patient), and a temporal transformer. The clinical notes are processed by a BioClinical BERT so as to extract meaningful representations. The resulting features from these pipelines for each modality are then concatenated and aggregated by a fusion transformer to generate hospital readmission predictions for each admission record.}
\label{overview}
\vspace{-0.3cm}
\end{figure}

\section{Methodology}
%
Fig.~\ref{overview} depicts our proposed MuST framework.
%
First, we employ clinical domain knowledge (i.e., $\phi_E$ in Fig.~\ref{overview}) derived from the hierarchical structure of medical codes \cite{tariq2021patient, tang2022multimodal} and clinical expertise \cite{tang2022multimodal} to preprocess the input EHR. 
In parallel, we extract visual representations from the chest radiographs using DenseNet121 \cite{huang2017densely} (i.e., $\phi_I$ in Fig.~\ref{overview}).
To exploit the spatial and temporal relationships within EHR and images, we create a relationship graph and feed the resulting data into a temporal transformer. 
Each clinical note file is segmented into several chunks, which are then processed using BioClinical BERT \cite{alsentzer2019publicly} to extract a sequence of representations. 
A fusion transformer model is adopted to aggregate information across three different modalities and pass the output to a multi-layer perceptron to obtain predictions of future readmission for each admission record. 

\subsection{Problem Formulation}

Hospital admissions can be treated as a sequence of visit records consisting of information from multiple modalities, such as patient EHR, medical images, and clinical note. For the visit record of the m-th patient $X_m$, it can be represented as $X_m=[x_{m,1}, x_{m,2}, ..., x_{m,t}, ..., x_{m,T_m}]$, where $T_m$ is the total number of visits for this patient in the given dataset, $t\in\{1,2,..,T_m\}$. $x_{m,t}$ can be further denoted as $x_{m,t} = \{e_{m,t}^o, i_{m,t}^o, n_{m,t}^o\}$, where $e_{m,t}^o \in E_m^o$, $i_{m,t}^o \in I_m^o$, $n_{m,t}^o \in N_m^o$, $E_m^o$, $I_m^o$, and $N_m^o$ are the original EHR, chest radiograph, and clinical note data for the m-th patient, respectively. Given the previous visits $[x_{m,1}, x_{m,2}, ..., x_{m,{(t-1)}}]$ of a discharged patient, the objective for hospital readmission prediction task is to predict whether the patient will be admitted again to the hospital within certain time span (e.g., 30 days in our case). 

\subsection{Feature Extraction for Single Modality}

\para{Graph Construction for Hospital Admissions.}
For the MIMIC-IV dataset as our original EHR input $E^o$, we choose to include demographics (age, gender, ethnicity), comorbidities (i.e., ICD-10 codes), lab tests, and medications. We follow the EHR preprocessing procedures \cite{tariq2021patient, tang2022multimodal} to incorporate clinical domain knowledge ($\phi_E$) consisting of both the hierarchical structure of medical codes and the clinical expertise. Meanwhile, we utilize a pre-trained DenseNet121 model ($\phi_I$) \cite{huang2017densely, he2020momentum, irvin2019chexpert} to perform image preprocessing on chest radiographs $I^o$ from MIMIC-CXR-JPG dataset. These can be represented as:
\begin{align}
V_E=\phi_E(E^o), \quad
V_I=\phi_I(I^o), \nonumber
\end{align}
where the resulting $V_E \in \mathbb{R}^{a\times l_{ehr}\times d}$ and $V_I \in \mathbb{R}^{a\times l_{img}\times d}$ are the node features for EHR and imaging modalities. 
$a$ is the total number of admissions, $l$ is the sequence length, and $d$ is the hidden dimension.

The edges of a graph depicting hospital admissions represent the degree of similarity between the features of medical notes for each modality. 
Patients with similar demographics and medical records may exhibit similarities in disease prognosis, diagnosis, and readmission probability. 
Therefore, the adjacency matrix $\mathcal{A} \in \mathbb{R}^{a\times a}$ can be represented as:
\begin{align}
 \mathcal{A}_{ij} = \begin{cases}
    exp\{-\frac{dist(v_i, v_j)^2}{\sigma^2}\}, & \text{if $\mathcal{A}_{ij}\geq \delta$}, \nonumber \\ 
    0, & \text{otherwise}.
  \end{cases}
\end{align}, 
where $v_i, v_j \in V_E$ or $V_I$; $\delta$ is designed to filter out the weak connections; $dist(v_i, v_j)$ is the distance/similarity between $v_i$ and $v_j$. A graph representation for hospital admissions can then be denoted as $\mathcal{G}_E = \{V_E, \mathcal{A}\}$ for EHR and $\mathcal{G}_I = \{V_I, \mathcal{A}\}$ for chest radiograph. 

Finally, to enable message passing for the graph representation of each modality, we adopt GraphSAGE \cite{hamilton2017inductive}:
\begin{align}
    \mathcal{G}_E^\mathcal{S} = \sigma(GraphSAGE(\mathcal{G}_E)), \quad
    \mathcal{G}_I^\mathcal{S} = \sigma(GraphSAGE(\mathcal{G}_I)), \nonumber
\end{align}
where $\mathcal{S}$ stands for "spatial"; $\sigma$ represents the activation function.

\para{Temporal Representation Learning.}
To capture the temporal dependency between each node feature, we propose to employ a temporal transformer ($\mathbf{T}_{temp}$) model. Each temporal transformer layer can be represented as
\begin{align}
    e_{l+1}^\mathcal{ST} = \mathbf{T}_{temp}(e_l^\mathcal{ST}; \theta_E), \quad
    i_{l+1}^\mathcal{ST} = \mathbf{T}_{temp}(i_l^\mathcal{ST}; \theta_I), \nonumber
\end{align}
where $\mathcal{ST}$ stands for "spatiotemporal"; $\theta_E$ and $\theta_I$ are model parameters that are different for each modality. Note that $e_0^{\mathcal{ST}} = V_E$ and $i_0^{\mathcal{ST}} = V_I$. We extract the final output representations of the \textit{CLS} token \cite{devlin2018bert}, $E^\mathcal{ST} = e_{L,cls}^\mathcal{ST}$ and $I^\mathcal{ST} = i_{L, cls}^\mathcal{ST}$, as the outputs of the overall \textit{ST-Transformers}.

\para{Feature Embedding for Clinical Notes.}
Clinical notes offer a comprehensive overview of the patient in comparison to structured features \cite{huang2019clinicalbert}. 
They encompass a wide range of information, including symptoms, diagnostic rationales, radiology findings, daily routines, and patient background \cite{huang2019clinicalbert}. 
We include unstructured clinical information to our pipeline by utilizing  MIMIC-IV-Note dataset, $N^o$. 
BioClinical BERT \cite{alsentzer2019publicly} is a variant of BERT \cite{devlin2018bert} model that has been pre-trained on clinical text to learn the domain-specific knowledge. 
We would like to use it for extracting meaningful features from the clinical notes in MIMIC-IV-Note to boost our model performance. 
However, according to \cite{devlin2018bert} and \cite{huang2019clinicalbert}, the maximum sequence length supported by the model is set to 512. To efficiently handle the extensive note files, it is necessary to divide them into chunks,  process each chunk individually using BioClinical BERT ($\phi_N$), and subsequently extract the \textit{CLS} token from each processed chunk as a representation \cite{devlin2018bert}:
\begin{align}
    N^{\mathcal{B}} = W_d[\phi_N^{CLS}(N_1^o), \phi_N^{CLS}(N_2^o), ..., \phi_N^{CLS}(N_c^o)], \nonumber
\end{align}
where $c$ is the total number of chunks. To account for potential variations in the number of chunks, we ensure that the list of chunk representations is padded to accommodate the maximum number of chunks, which is $C=25$. $\mathcal{B}$ stands for "BERT", $N^{\mathcal{B}} \in \mathbb{R}^{a\times C \times d}$. $W_d$ is a projection matrix that converts the hidden dimension to $d$, aligning it with the dimensions of the other two modalities to facilitate subsequent fusion.

\subsection{Multimodal Fusion}

The final step of our pipeline involves combining the spatiotemporal features $E^\mathcal{ST}$ and $I^\mathcal{ST}$ for EHR and chest radiograph with the clinical note representations $N^{\mathcal{B}}$. We propose to employ a fusion transformer model ($\mathbf{T}_{fusion}$) for the purpose of aggregating information across all three modalities. Let
\begin{align}
    \mathbf{Z} = E^\mathcal{ST}\oplus I^\mathcal{ST}\oplus N^{\mathcal{B}}, \nonumber
\end{align}
where $\oplus$ is the sign for concatenation and $\mathbf{Z} \in \mathbb{R}^{a\times (l_{ehr}+l_{img}+C) \times d}$. Denote a multimodal transformer layer, $\mathbf{Z}_{l+1} = \mathbf{T}_{fusion}(\mathbf{Z}_{l})$ as
\begin{align}
    \mathbf{y}_l = \mathrm{MSA}(\mathrm{LN}(\mathbf{Z}_l))+\mathbf{Z}_l, \quad
    \mathbf{Z}_{l+1} = \mathrm{MLP}(\mathrm{LN}(\mathbf{y}_l))+\mathbf{y}_l, \nonumber
\end{align}
where LN is Layer Normalization, $\mathrm{MLP}$ is Multilayer Perceptron, and MSA is Multi-Headed Self-Attention. Let $E_{l, norm}^\mathcal{ST}$, $I_{l, norm}^\mathcal{ST}$, and $N_{l, norm}^\mathcal{B}$ denote the representations of EHR, chest radiographs, and clinical notes at layer $l$ after LN operation, respectively. Then $\mathrm{MSA}(\mathbf{Z}_{l,norm}))$ can be represented as
\begin{align}
    \mathrm{MSA}(\mathbf{Z}_{l,norm}))&=\mathrm{Attention}(\mathbf{W}^Q\mathbf{Z}_{l,norm}, \mathbf{W}^K\mathbf{Z}_{l,norm}, \mathbf{W}^V\mathbf{Z}_{l,norm}), \nonumber\\
    &=(\bigoplus_{j=1}^J \mathbf{H}_j)W_o, \nonumber
\end{align}
where
\begin{align}
    \mathbf{H}_j&=softmax(\frac{(\mathbf{Z}_{l,norm}W_j^Q)(\mathbf{Z}_{l,norm}W_j^K)^\intercal}{\sqrt{d_K}}\mathbf{Z}_{l,norm}W_j^V), \nonumber\\
    \mathbf{Z}_{l,norm}&=E_{l, norm}^\mathcal{ST}\oplus I_{l, norm}^\mathcal{ST}\oplus N_{l, norm}^\mathcal{ST}, \nonumber
\end{align}
$\mathbf{W}^Q$, $\mathbf{W}^K$, $\mathbf{W}^V$, and $W_o$ are projection matrices. The MSA mechanism enables the fusion transformer model to automatically identify and assign greater weight to the most influential modality, thereby enhancing the final prediction.

Finally, we obtain an output representation that is of the same dimension as $\mathbf{Z}$. We perform $Max()$ operation to the second dimension and pass the output to an MLP to generate the final hospital readmission predictions. The objective function is set to be the binary cross entropy $\mathcal{L}_{bce}$.


\section{Experiments}
\para{Datasets.}
We assess the efficacy of the proposed MuST framework on the MIMIC dataset, which is a combination of three datasets: Medical Information Mart for Intensive Care IV (MIMIC-IV) v1.0\cite{Johnson2023, physiobank} dataset for EHR, MIMIC-CXR-JPG v2.0.0\cite{johnson2019mimic, physiobank} dataset for chest radiographs, and MIMIC-IV-Note \cite{johnson2023mimicnotes, physiobank} dataset for clinical notes.  
The samples in the three datasets have a unique ID to link together. The following provides a more detailed description of the datasets.

\begin{itemize}
    \item \textbf{MIMIC-IV dataset.} MIMIC-IV is a collaborative effort between Beth Israel Deaconess Medical Center (BIDMC) and Massachusetts Institute of Technology (MIT). The study included 431,231 hospital admissions from 180,733 distinct patients.

    \item \textbf{MIMIC-CXR-JPG dataset.} MIMIC-CXR dataset comprises 227,835 imaging studies conducted on 64,588 patients who sought treatment at the BIDMC Emergency Department from 2011 to 2016. The dataset contains 377,110 images.

    \item \textbf{MIMIC-IV-Note dataset.} MIMIC-IV-Note dataset consists of 331,794 deidentified discharge summaries from 145,915 patients who were admitted to the hospital and emergency department at BIDMC. 
\end{itemize}
In our study, we first follow the criteria adopted in \cite{tang2022multimodal} for subsetting the MIMIC-IV and MIMIC-CXR-JPG datasets. Since the clinical notes are also incorporated into the proposed pipeline, we obtain the intersection of admissions from the previous subset and the MIMIC-IV-Note dataset. The resulting dataset consists of 13,763 hospital admissions, with a distribution of 8,772 for training, 2,777 for testing, and 2,214 for validation. It includes a total of 82,465 chest radiographs and 82,465 clinical notes from 11,041 different patients. Among these patients, 2,379 admissions involved readmission within 30 days of discharge.

\begin{table}[t]
\centering
\caption{Comparisons with past research. Following \cite{tang2022multimodal}, confidence intervals (CI) were calculated using the Delong method \cite{delong1988comparing}. The best results are highlighted in bold. The subscript "coatt" stands for the case where an extra co-attention mechanism is added prior to fusion. Subscripts "mean", "last", and "max" stand for different operations to pool the temporal features. *: our approach.}
\begin{tabular}{m{2.50cm}<{\centering}m{3.00cm}<{\centering}m{3.20cm}<{\centering}m{2.50cm}<{\centering}}
\toprule[1pt]
\multirow{2}{*}{Model} & \multirow{2}{*}{Modality} & \multicolumn{2}{c}{MIMIC-IV}\\ 
\cmidrule(r){3-4}&  & AUC (\%) [95\% CI] & ACC (\%)\\
\bottomrule[1pt]


STGNN  &
Image &
71.86 [69.37 74.34]  &
74.46\\

STGNN &
EHR &
78.32 [75.89 80.75] &
85.32\\

MM-STGNN\textsubscript{coatt} &
EHR+Image &
76.74 [74.19 79.28] &
85.29\\

MM-STGNN\textsubscript{mean} &
EHR+Image &
79.15 [76.82 81.48] &
85.63\\

MM-STGNN\textsubscript{last} &
EHR+Image &
79.31 [76.91 81.71] &
85.87\\

MM-STGNN\textsubscript{max} &
EHR+Image &
79.46 [77.11 81.81] &
85.31\\

\bottomrule[1pt]
ST-Transformer* &
Image &
72.07 [69.59 74.54] &
74.35\\

ST-Transformer* &
EHR &
78.42 [75.97 80.87] &
85.01\\

CN-Transformer* &
Text &
83.06  [80.67 85.45] &
91.75\\

MuST* &
EHR+Image+Text &
\textbf{85.81 [83.66 87.97] } &
\textbf{92.37} \\
\bottomrule[1pt]
\end{tabular}
\label{results}
\vspace{-0.3cm}
\end{table}

\para{Experimental Setup and Evaluation Metrics.} 
Experiments were conducted using the Adam optimizer \cite{kingma2014adam} in PyTorch \cite{paszke2019pytorch} on a workstation with eight NVIDIA GeForce RTX 3090 (24 GB) GPUs. 
The final hyperparameters for our proposed MuST model were as follows: (a) learning rate was 0.001; (b) dropout was 0.1; (c) number of epochs was 300; (d) maximum sequence length for both EHR and chest radiographs was 9; (e) number of layers for two temporal transformers and a fusion transformer was 1.
The accuracy (ACC) and area under the curve (AUC) are used as the evaluation metrics. 
%


\para{Comparison with Previous Methods.}
We compare our proposed MuST framework with the previous methods STGNN and MM-STGNN \cite{tang2022multimodal} for 30-day all-cause hospital readmission prediction task performed on the MIMIC-IV dataset in Table~\ref{results}.
Our model demonstrates satisfactory performance in both AUC and ACC for predicting hospital readmission. 
With a single CN-Transformer (ClinicalNote-Transformer, which is constructed by passing the clinical note features extracted by BioClinical BERT to a transformer), we outperform MM-STGNN by almost 4\% on AUC and 6\% on ACC.
The integration of EHR, medical imaging, and textual data results in an overall performance of of 85.81\% AUC and 92.37\% ACC.
In conclusion, our model greatly benefits from its multimodal nature and outperforms the previous state-of-the-art.

\para{Ablation Analysis.}
We further conduct an ablation study and demonstrate the results in Table~\ref{ablation}. 
We test the effect of: (1) temporal layers (4th\&5th rows); (2) fusion layers (2nd\&5th rows); (3) textual data (1st\&2nd and 3rd\&5th rows).
Refer to Fig.~\ref{overview} for the definition of the ST-Transformer.
The ablation analysis results show that all of the components in MuST contribute to the performance of the model to a certain extent. 
Obviously, losing the text data significantly decreases the model's performance. 
Furthermore, for the MuST model specifically, fusion transformer layers have a greater impact on the final predictions than temporal transformer layers.


\begin{table}[t]
\centering
\caption{Ablation analysis. Best results are highlighted in bold.}
\begin{tabular}{m{3.00cm}<{\centering}m{3.00cm}<{\centering}m{3.20cm}<{\centering}m{2.50cm}<{\centering}}
\toprule[1pt]
\multirow{2}{*}{Model} & \multirow{2}{*}{Modality} & \multicolumn{2}{c}{MIMIC-IV}\\ 
\cmidrule(r){3-4}&  & AUC (\%) [95\% CI] & ACC (\%)\\
\bottomrule[1pt]
ST-Transformer + MLP&
EHR+Image &
78.88 [76.46 81.29] &
86.48\\

ST-Transformer + MLP&
EHR+Image+Text &
83.83 [81.50 86.17] &
92.22\\

MuST (w/o clinical notes)&
EHR+Image &
79.87  [77.52 82.23] &
85.87\\

MuST (w/o temporal layers)&
EHR+Image+Text &
84.23 [81.90 86.56] &
91.93\\

\bottomrule[1pt]
MuST &
EHR+Image+Text &
\textbf{85.81 [83.66 87.97] } &
\textbf{92.37} \\
\bottomrule[1pt]
\end{tabular}
\label{ablation}
\vspace{-0.3cm}
\end{table}
\section{Conclusion}
This paper introduces MuST, a framework that efficiently utilizes the multimodal and spatiotemporal properties of clinical data for hospital readmission prediction.
We extract the representations from different data modalities with tailored network architectures and then adopt a generic fusion transformer model to aggregate information across three different modalities and pass the output to a multi-layer perceptron to obtain predictions of future readmission for each admission record. 
Extensive experiments on the MIMIC-IV dataset show that our proposed framework is both effective and efficient, with promising results. 
For future studies, we may evaluate our model on more internal and external datasets to validate the generalizability of the pipeline and also investigate other fusion approaches to improve the efficiency of our framework.

\para{Acknowledgments.} The work described in this paper was partially supported by grants from the National Natural Science Foundation of China (No. 62201483), the Research Grants Council of the Hong Kong Special Administrative Region, China (T45-401/22-N) and HKU seed fund for basic research (202111159073).

\clearpage
\bibliographystyle{splncs04}
\bibliography{reference}

%




\end{document}